\documentclass[sigconf]{aamas} 
\usepackage{balance} 

\usepackage{amsmath,amssymb,mathtools,bm}
\usepackage{algorithm}
\usepackage{algpseudocode} 
\usepackage{microtype}
\usepackage{xspace}

\newcommand{\X}{\mathcal{X}}
\newcommand{\R}{\mathbb{R}}
\newcommand{\Nc}{\mathcal{N}}
\newcommand{\Vol}{\mathrm{Vol}}
\newcommand{\cD}{\mathcal{D}}
\newcommand{\cS}{\mathcal{S}}
\newcommand{\E}{\mathbb{E}}
\newcommand{\ind}[1]{\mathbf{1}\{#1\}}
\newcommand{\inds}[2]{\mathbf{1}_{#1}\!\left(#2\right)}
\newcommand{\norm}[1]{\left\lVert #1 \right\rVert}
\newcommand{\gp}{\mathcal{GP}}
\newcommand{\Ball}[2]{B_{#1}\!\left(#2\right)}
\newcommand{\MOCCAS}{{\textsc{moc-cas}}\xspace}

\doi{XTVI9400}

\makeatletter
\gdef\@copyrightpermission{
  \begin{minipage}{0.2\columnwidth}
   \href{https://creativecommons.org/licenses/by/4.0/}{\includegraphics[width=0.90\textwidth]{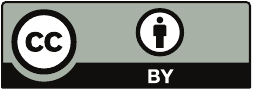}}
  \end{minipage}\hfill
  \begin{minipage}{0.8\columnwidth}
   \href{https://creativecommons.org/licenses/by/4.0/}{This work is licensed under a Creative Commons Attribution International 4.0 License.}
  \end{minipage}
  \vspace{5pt}
}
\makeatother

\setcopyright{ifaamas}
\acmConference[AAMAS '26]{Proc.\@ of the 25th International Conference
on Autonomous Agents and Multiagent Systems (AAMAS 2026)}{May 25 -- 29, 2026}
{Paphos, Cyprus}{C.~Amato, L.~Dennis, V.~Mascardi, J.~Thangarajah (eds.)}
\copyrightyear{2026}
\acmYear{2026}
\acmDOI{}
\acmPrice{}
\acmISBN{}

\title[AAMAS-2026 Formatting Instructions]{Multi-Objective Coverage via Constraint Active Search}

\author{Zakaria Shams Siam}
\affiliation{
  \institution{University at Albany, State University of New York}
  \city{Albany}
  \country{United States}}
\email{zsiam@albany.edu}

\author{Xuefeng Liu}
\affiliation{
  \institution{University of Chicago}
  \city{Chicago}
  \country{United States}}
\email{xuefeng@uchicago.edu}

\author{Chong Liu}
\affiliation{
  \institution{University at Albany, State University of New York}
  \city{Albany}
  \country{United States}}
\email{cliu24@albany.edu}

\begin{abstract}
In this paper, we formulate the new multi-objective coverage (MOC) problem where our goal is to identify a small set of representative samples whose predicted outcomes broadly cover the feasible multi-objective space. This problem is of great importance in many critical real-world applications, e.g., drug discovery and materials design, as this representative set can be evaluated much faster than the whole feasible set, thus significantly accelerating the scientific discovery process. Existing works cannot be directly applied as they either focus on sample space coverage or multi-objective optimization that targets the Pareto front. However, chemically diverse samples often yield identical objective profiles, and safety constraints are usually defined on the objectives. To solve this MOC problem, we propose a novel search algorithm, MOC-CAS, which employs an upper confidence bound-based acquisition function to select optimistic samples guided by Gaussian process posterior predictions. For enabling efficient optimization, we develop a smoothed relaxation of the hard feasibility test and derive an approximate optimizer. Compared to the competitive baselines, we show that our MOC-CAS empirically achieves superior performances across large-scale protein-target datasets for SARS-CoV-2 and cancer, each assessed on five objectives derived from SMILES-based features.
\end{abstract}

         
\newcommand{\BibTeX}{\rm B\kern-.05em{\sc i\kern-.025em b}\kern-.08em\TeX}

\begin{document}

\pagestyle{fancy}
\fancyhead{}

\maketitle 

\section{Introduction}

\begin{figure}[t]
    \centering
    \begin{minipage}{0.6\linewidth}\centering
		\includegraphics[width=\textwidth]{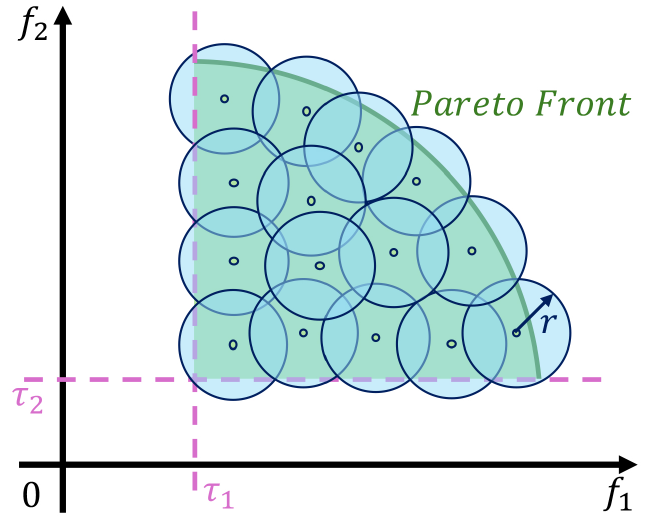}
	\end{minipage}
    \caption{An example of the MOC problem where $f_1,f_2$ are two objective functions and $\tau_1, \tau_2$ are two thresholds defining the feasible region shown in light green. 15 samples are selected as representative feasible samples, each with a coverage ball whose radius is $r$.}
    \label{fig:moc}
    \Description{Illustration of a multi-objective coverage problem in two-dimensional objective space. Two threshold lines define a feasible region. Fifteen feasible samples are plotted inside the feasible region, and each sample is surrounded by a circular coverage neighborhood with radius r to show how the selected samples collectively cover the feasible region.}
\end{figure}

Many critical real-world applications, such as drug discovery and materials design, increasingly rely on data-driven screening to identify a small set of promising designs for expensive downstream analysis. Often, the acceptance of candidate designs is governed by a feasible region defined by per-objective lower bounds or acceptance thresholds (e.g., for docking, solubility, or synthetic accessibility). Under limited experimental budgets, procedures are therefore sought that yield a compact yet diverse set of designs satisfying these acceptability criteria while requiring as few costly experiments as possible \citep{li2025none}. This motivates active and sample-efficient search algorithms that reason directly about the specific outcomes which will be most informative and cost-effective for the decision makers \citep{snoek2012practical,frazier2018tutorial}.

Classical multi-objective optimization methods typically emphasize approximating the Pareto frontier and improving hypervolume \citep{knowles2006parego,emmerich2011hypervolume,yang2019efficient,li2025constrained}. These objectives are well suited for managing trade-offs among competing objectives, but they do not directly target the discovery of threshold-compliant candidates. Contemporary works on \emph{constraint active search} (CAS) contribute toward discovery under acceptance/rejection criteria, however, the state-of-the-art \citep{malkomes2021beyond} frames coverage in the sample space and applies expected improvement (EI)-like utilities. However, sample-space coverage can be misaligned with many decision criteria (e.g., in molecular design), where chemically diverse inputs may map to similar objective profiles. In such settings, input-space novelty can waste evaluations on designs that look different yet yield nearly identical outcomes, while failing to cover the feasible region where decisions are made: the objective space.

In contrast, coverage in the output (objective) space aligns directly with downstream selection, as acceptability is inherently defined there. Optimizing output-space coverage therefore prioritizes the discovery of a few representative outcome patterns that satisfy all imposed constraints, enabling faster and more reliable down-selection than strategies that only spread samples in the input space or trace the Pareto front. Our goal is to select a compact set of samples whose outcomes are well distributed across the feasible region, rather than merely spreading inputs in sample space or tracing the Pareto frontier. This supports reliable down-selection and risk-aware decision making under uncertainty. See Figure \ref{fig:moc} for an example with two objective functions.

To solve this new and highly motivated problem, we formulate the new \emph{multi-objective coverage} (MOC) problem, where per-objective thresholds define a feasible region in objective space. The goal is to identify a small set of diverse, representative samples whose outcomes satisfy the thresholds and collectively cover the feasible region under uncertainty—without attempting to enumerate all feasible points \citep{gotovos2013active}, which can be prohibitively expensive. This new problem yields a compact and diverse panel that supports decision-making under uncertainty while minimizing the number of costly evaluations. To solve the MOC problem, We model each objective with an independent Gaussian process (GP) \citep{williams2006gaussian} and design an optimistic acquisition function that trades off feasibility and incremental objective-space coverage gain, using upper confidence bounds (UCB) \citep{srinivas2010gaussian}.

\noindent\textbf{Contributions.} Our contributions are summarized as follows.
\begin{itemize}
    \item We formulate the new MOC problem, which explicitly targets fast coverage and diversity of the feasible objective region under per-objective thresholds. This is distinct from the CAS framed in the sample space \citep{malkomes2021beyond} and also different from other frontier-centric objectives \citep{knowles2006parego,emmerich2011hypervolume,yang2019efficient}.
    \item To solve the problem, we propose the \emph{multi-objective coverage via constraint active search} (MOC-CAS) algorithm, which is a novel constrained active search algorithm that evaluates each candidate based on the optimistic estimate of the new feasible volume it is expected to cover within the feasible region. It further employs a tie-breaking mechanism that promotes diversity by encouraging dispersion in the predicted objective values.
    \item To enable efficient optimization, we replace the hard feasibility indicators with smooth surrogates and thus derive an approximate optimizer for the proposed acquisition function.
    \item On large-scale protein-target datasets of 1 million compounds spanning cancer and SARS-CoV-2 \citep{liu2023drugimprover}, our MOC-CAS algorithm achieves better performance than competitive baselines under similar evaluation budgets.
\end{itemize}

\noindent\textbf{Technical Novelties.} Our technical novelties are:

\begin{itemize}
    \item Our \MOCCAS selects samples by \emph{maximizing newly uncovered neighborhood volume} in the objective space at a fixed coverage resolution, utilizing a UCB prediction to estimate feasibility before any coverage value is assigned. To maintain diversity in the objective space, we break ties by selecting the candidate with the greatest objective-space distance to prior feasible outcomes. 
    \item We derive a mass-preserving relaxation of the geometric objective with hard indicators. We replace the ball indicator by a unit-mass Gaussian kernel. Additionally, we substitute a smooth per-objective probit gate for the orthant indicator, and further replace the union of covered balls by a soft kernel sum. The resulting local average therefore yields a closed-form novelty term and a fully differentiable acquisition function along with explicit gradients via the chain rule, thus enabling fast optimization in practice.
\end{itemize}

\section{Related Work}\label{sec:rw}

We briefly discuss related work from three different perspectives.

\subsection{Constraint Active Search}

Constraint active search (CAS) reexamines multiobjective design from Pareto-frontier tracing for finding a diverse set of threshold-satisfying samples. Past work in \citep{malkomes2021beyond} presented CAS with the expected coverage improvement (ECI) acquisition function that scores samples by their expected increase in the covered feasible volume in the sample space, where where coverage is measured in the sample space, using fill distance \citep{malkomes2021beyond} as a proxy for diversity. CAS has also been utilized for optimizing performance during machine learning model deployment by connecting offline and online experimentations \citep{komiyama2022bridging}. A related but distinct line of work \citep{misra2022learning} learns active constraint sets in order to speed up repeated deterministic optimization by estimating which constraints bind at optimality where the objective is to find faster solution for parametric programs with theoretical guarantees, instead of applying probabilistic search for finding diverse satisfactory samples. In contrast, our proposed approach explicitly defines and directly optimizes coverage in the objective space under multi-threshold feasibility. It introduces an optimism-based feasibility gate linked to incremental objective-space coverage, thereby reshaping the acquisition geometry and the mechanism by which diversity is enforced, while preserving the core intuition of CAS methods - that a few well-chosen samples can effectively cover the feasible set.

\subsection{Level Set Estimation}

The level set estimation (LSE) problem was first approached in \citep{bryan2005active} through the straddle heuristic algorithm by utilizing a GP-based active learning technique. Subsequent work in \citep{gotovos2013active} presented the LSE algorithm for active level-set estimation along with Gaussian-process confidence bounds, derived sample-complexity guarantees, and further extended the method to implicit thresholds and also batched sampling.
Also, TRUVAR, a unified Bayesian framework for Bayesian optimization (BO) and LSE was proposed in \citep{bogunovic2016truncated} which greedily diminishes the truncated posterior variance over sample sets, controls pointwise costs and heteroscedastic noise, and also comes with theoretical guarantees. Zanette et al. \citep{zanette2018robust} maximized the expected raise in the volume of the estimated level set. Wang et al. \citep{wang2019optimization} developed a local–minimax framework for noisy zeroth-order optimization and proposed that instance-dependent rates could be substantially faster when the reference function’s level sets satisfy fast volume-growth conditions and regularity, explicitly showing that the problem’s difficulty is directly related to and also determined by the properties of the level-set geometry.

Iwazaki et al. \citep{iwazaki2020bayesian} focused on the LSE problem with input uncertainty where the input parameters are vulnerable to perturbation from a known Normal distribution. This study is closely related to safe BO research \citep{sui2015safe, sui2018stagewise}, where they sample only the configurations that surpass the threshold with high probability. Subsequent study in \citep{mason2022nearly} proposed nearly-optimal (instance-dependent and non-asymptotic) sample-complexity bounds for explicit and implicit level-set estimation by connecting the problem to adaptive experimental design in reproducing kernel Hilbert space (RKHS), with methods which match known lower bounds in linear (kernel) settings.

\subsection{Multi-Objective Optimization} 

Multi-objective optimization (MOO) aims to effectively approximate the Pareto front under several competing objectives. Early probabilistic techniques utilized estimation-of-distribution ideas to design Bayesian model-based evolutionary algorithms which maintain diverse approximations to the Pareto front \citep{laumanns2002bayesian}. The work in \citep{laumanns2002bayesian} combines a BO algorithm into a multi-objective evolutionary algorithm and investigate its behavior on combinatorial problems, highlighting the role of probabilistic modeling for convergence and diversity. 
Information-theoretic acquisition functions in Bayesian optimization, such as, PESMO select evaluations in order to maximally reduce the entropy of the posterior over the Pareto set and allow a decomposition across all objectives, thus enabling decoupled evaluation with computational cost which scales linearly in the number of objectives \citep{hernandez2016predictive}. 

Another line of work utilizes random scalarizations where the authors proposed a flexible framework which samples scalarization weights to the target specified regions of the Pareto front and derive sublinear regret guarantees along with favorable computational cost \citep{paria2020flexible}. 
Complementarily, it was also shown that dominated hypervolume could be written as an expectation over hypervolume scalarizations, producing algorithms with hypervolume regret bounds when paired with standard BO (UCB or Thompson sampling) methods and a simple predictor for hypervolume \citep{zhang2020random}. 
Recent scalable MOO-BO approaches develop parallel expected hypervolume improvement (EHVI) and trust-region techniques for higher dimensions and large batches (e.g., qEHVI/qNEHVI and MORBO) \citep{daulton2021parallel, daulton2022multi, li2025constrained}. Park et al. \citep{park2023botied} proposed BOtied which is based on a cumulative distribution function indicator. 
Unlike these Pareto front-centric approaches (e.g., hypervolume, entropy, or scalarization driven), our proposed work seeks coverage in objective space under explicit multi-threshold feasibility by optimizing a coverage-gain objective which rewards discovering diverse feasible outcomes instead of solely tracing the Pareto front.

\section{Preliminaries}\label{sec:pre}

\subsection{Problem Statement}\label{sec:pre:problem}

Let $\X\subseteq \R^d$ be a compact sample space and
$f=(f_1,\dots,f_m):\X\to\R^m$ a black-box, unknown, vector-valued function with $m$ objectives.
A noisy query at $x\in\X$ gives $y_i=f_i(x)+\varepsilon_i$,
where $\varepsilon_i\sim\Nc(0,\sigma^2)$ are i.i.d. Gaussian noise terms, for $i=1,\dots,m$.
Given per-objective thresholds $\tau=(\tau_1,\dots,\tau_m)\in\R^m$, an outcome is called \emph{feasible} provided that all objectives meet their thresholds.
This defines the feasible region in the objective space:
\[
\cS \;=\; \bigl\{\bm z\in\R^m:\; z_i\ge\tau_i\ \ \forall i\bigr\}.
\]
In many critical real-world applications, e.g., drug discovery and materials design, the decision criterion is defined on the objective values, not molecular inputs. Therefore, covering representative outcomes inside $\cS$ is more consistent and aligned with down-selection than finding the Pareto frontier.

We select designs $x_1,x_2,\dots \in\X$ sequentially and observe $y_t=f(x_t)+\varepsilon_t$.
At the end of $t$ steps, $\cD_t=\{(x_s,y_s)\}_{s=1}^t$ and the observed outcomes are $Z_t=\{y_s\}_{s=1}^t\subset\R^m$.
We fix a \emph{resolution} parameter $r>0$. In real-world problem settings, we can define $r$ with respect to the robustness of the design to perturbations. Many experimental design problems typically have a sense of known resolution (e.g., manufacturing precision/tolerance or simulation accuracy). We interpret $r$ as a coverage resolution: outcomes within distance $r$ are treated as effectively redundant for representing $\cS$. For $r>0$ and center $\bm z\in\R^m$,
\begin{align*}
    \Ball{r}{\bm z}=\{\bm z'\in\R^m:\norm{\bm z'-\bm z}_2<r\}
\end{align*}
represents the coverage neighborhood of $\bm z$ which is an $r$-ball in the output space. We thus define the covered volume for a set of points $Z_t$ as
\[
u_t \;=\; \Vol \Big(\,\Ball{r}{Z_t}\cap\cS \Big),
\qquad 
\Ball{r}{Z_t} \;=\; \bigcup_{\bm z\in Z_t} \Ball{r}{\bm z}.
\]
Our objective is to \emph{maximize $u_t$ with as few evaluations as possible}, thus generating a small and diverse set of threshold-compliant or feasible outcomes that broadly covers $\cS$ at resolution $r$.

\noindent\textbf{Evaluations.}
No single metric fully captures performance in multi-objective coverage. We therefore report three complementary metrics.

\textsc{Fill Distance.} Fill distance quantifies how evenly the sampled objective vectors cover the feasible region in objective space. The fill distance is formally defined as follows where $d(\cdot,\cdot)$ denotes the Euclidean distance in the objective space. 
\begin{equation*}
\mathrm{FILL}(Z_t,\cS)
\;=\;
\sup_{\mathbf{z}\in \cS}\; \min_{\mathbf{z}_j\in Z_t} d \left(\mathbf{z}_j,\mathbf{z}\right).
\end{equation*}
$\mathrm{FILL}(Z_t,\cS)$ represents the radius of the largest empty ball that one can fit in $\cS$ in Euclidean space \citep{malkomes2021beyond}. It measures the spacing of $Z_t$ in $\cS$. If a set’s fill distance is smaller with respect to $\cS$, then that set is better distributed within $\cS$.

\textsc{Positive Samples.} This represents the number of feasible points in the objective space (inside the feasible set, $\cS$) for the samples selected by the algorithm.

\textsc{Area Under the Positives Curve (AUP).} We further summarize sample efficiency utilizing the area under the cumulative discoveries curve. For cumulative positives $P(t)$ up to round $t$, we calculate
\[
\mathrm{AUP} \;=\; \sum_{t=1}^{T} P(t),
\]
or its trapezoidal variant. AUP rewards those methods which find good samples both \emph{early} and \emph{consistently} across the budget. Higher AUP indicates better performance by the method (since earlier and steadier discoveries score higher in this case). This metric is analogous to the area under the learning curve metric which is widely utilized in active learning research \citep{guyon2011results}.

\noindent\textbf{Additional Notations.}
We denote the scalars as italics (e.g., $t,r$), vectors as bold (e.g., $\bm z$), and sets as calligraphic (e.g., $\X,\cS$).
The Euclidean norm is denoted as $\norm{\cdot}$. $\Vol(\cdot)$ represents the Lebesgue volume in $\R^m$.
We use $\ind{\cdot}$ to represent an indicator where $\ind{x}=1$ if the condition $x$ holds and $0$ otherwise. Similarly, $\inds{\cdot}{\cdot}$ represents a set–membership indicator where $\inds{A}{z}=1$ if $z\in A$ and $0$ if $z\notin A$. We use $\gp$ for Gaussian processes.

\subsection{Background}\label{sec:pre:background}

\noindent\textbf{Constraint Active Search.}
Say $\mathcal{X}$ is a discrete or finite sample set having unknown, black-box multi-objective response $\tilde y:\mathcal{X} \to\mathbb{R}^m$. For the given collections of per-objective minimum thresholds, $\tau\in\mathbb{R}^m$, constraint active search (CAS) targets the feasible subset of the samples
\[
\mathcal{S}_\mathcal{X} \;=\; \{\,x\in\mathcal{X} : \tilde y_i(x)\ge \tau_i\ \ \forall i\in[m]\,\},
\]
and searches adaptively to identify a diverse set of feasible samples.

\noindent\textbf{Level Set Estimation.}
The objective of level set estimation (LSE) is to determine the region \(\{x\in\X:\ f(x)\ge \tau\}\) by sequentially querying \(x\) and then observing \(y=f(x)+\varepsilon\) given a (noisy) black-box objective \(f:\X\to\R\) and a threshold \(\tau\in\R\). With a GP posterior at round \(t-1\), let \(\mu_{t-1}(x)\) and \(\sigma_{t-1}(x)\) represent the mean and standard deviation, a typical substitute for “being above the level” is
\[
p_{t-1}(x)\;=\;\Pr \big(f(x)\ge\tau\mid \cD_{t-1}\big)\;=\;\Phi \Big(\frac{\mu_{t-1}(x)-\tau}{\sigma_{t-1}(x)}\Big),
\]
and the classic STRADDLE score selects points having high uncertainty near the level,
\[
s_{\text{straddle}}(x)\;=\;\kappa\,\sigma_{t-1}(x)\;-\;\big|\mu_{t-1}(x)-\tau\big|,
\]
mostly with \(\kappa\approx 1.96\). LSE methods are inclined to concentrate queries close to the decision boundary that separates \(\{f(x)\ge\tau\}\) from its complement, instead of sampling the interior of the superlevel set.

\noindent\textbf{Multi-Objective Optimization.}
Let \(f(x)=(f_1(x),\ldots,f_m(x)): \mathcal{X} \rightarrow \R^m\) be a function that needs to be maximized over a sample space \(\X\). For \(x_1,x_2\in\X\), we say \(x_1\) Pareto-dominates \(x_2\) if \(f_i(x_1)\ge f_i(x_2)\) for all \(i\) and \(f_j(x_1)>f_j(x_2)\) for some \(j\). The \emph{Pareto set} \(X^\star\subseteq\X\) accommodates all non-dominated samples and its image \(P=\{f(x):x\in X^\star\}\) is the \emph{Pareto front}. A typical scalar summary is the (reference-point) hypervolume indicator
\begin{align*}
&\quad \mathrm{HV}_{z}(Y)\\
&= \operatorname{vol} \Big(\,\big\{\,y\in\R^m:\ y\ge z,\ \text{and}\ \exists\,y'\in Y\ \text{with}\ y'\ \text{dominates}\ y\,\big\}\Big),
\end{align*}
for a finite set \(Y\subset\R^m\) of achieved outcomes and a reference point \(z\in\R^m\). Hypervolume enlarges as \(Y\) inflates the dominated portion of objective space beyond \(z\). It is widely utilized to compare methods which trade off multiple objectives.

\section{The MOC-CAS Algorithm}\label{sec:moccas}

In this section, we show full details of our \MOCCAS algorithm, summarized in Algorithm~\ref{alg:moccas}. Overall, \MOCCAS{} performs sequential experimental design in order to identify a diverse set of outcomes which satisfy all per–objective thresholds. At each round, we fit independent GPs for all the $m$ objectives, then form optimistic predictions, score each sample by an objective space coverage gain acquisition function which rewards new feasible coverage, and finally select the maximizer, while breaking ties in order to preserve objective–space diversity.

\noindent\textbf{Gaussian Process Modeling.}
We develop independent GP priors $f_i\sim\gp(0,k_i)$, $i=1,\dots,m$, to model all the objectives. Here, we refer to \(k_i(\cdot,\cdot)\) as the GP covariance function.
Given $\cD_{t-1}$, the posterior mean and variance at $x$ are denoted as $\mu^{(i)}_{t-1}(x)$ and $\sigma^{(i)}_{t-1}(x)^2$, shown as 
\begin{align*}
\mu^{(i)}_{t-1}(x)
&= \bigl(k^{(i)}_{t-1}(x)\bigr)^\top  \bigl(K^{(i)}_{t-1}+\sigma^2 I\bigr)^{-1}\mathbf y^{(i)}_{1:t-1}, \label{eq:gp-mean-i}\\
\sigma^{(i)}_{t-1}(x)^2
&= k_i(x,x) \;-\; \bigl(k^{(i)}_{t-1}(x)\bigr)^\top  \bigl(K^{(i)}_{t-1}+\sigma^2 I\bigr)^{-1} k^{(i)}_{t-1}(x),
\end{align*}
where $X_{t-1}=\{x_1,\ldots,x_{t-1}\}$ are the sample points queried so far and
$\mathbf y^{(i)}_{1:t-1}=[\,y^{(i)}_1,\ldots,y^{(i)}_{t-1}\,]^\top$ are the observations for objective $i$ with independent and identically distributed Gaussian noise variance $\sigma^2$, and $K^{(i)}_{t-1} \;=\; \bigl[k_i(x_a,x_b)\bigr]_{a,b=1}^{t-1}$, $k^{(i)}_{t-1}(x) \;=\; \bigl[k_i(x_1,x),\ldots,k_i(x_{t-1},x)\bigr]^\top$.

Using a confidence schedule $\beta_t>0$, we now define the optimistic UCB prediction
\begin{align*}
U^{(i)}_{t-1}(x) &= \mu^{(i)}_{t-1}(x) + \sqrt{\beta_t}\,\sigma^{(i)}_{t-1}(x),\quad
U_{t-1}(x) \vcentcolon= \bigl(U^{(i)}_{t-1}(x)\bigr)_{i=1}^{m}.
\end{align*}

\begin{algorithm}[t]
\caption{\textsc{MOC-CAS} Algorithm}
\label{alg:moccas}
\begin{algorithmic}[1]
\Require{GP prior kernels $\{k_i\}_{i=1}^m$, thresholds $\tau$, radius $r$, confidence schedule $(\beta_t)_{t\ge1}$, initial data $\cD_0$}
\For{$t=1,2,\dots,T$}
  \State Update independent GP posteriors using $\cD_{t-1}$
  \State $x_t \gets \arg\max_{x\in\X}\,\alpha^{\mathrm{\MOCCAS}}_{t-1}(x)$ \\
  \Comment{break ties by farthest $U_{t-1}(x)$ from $Z_{t-1}$}
  \State Query $y_t=f(x_t)+\varepsilon_t$
  \State Update $\cD_t=\cD_{t-1}\cup\{(x_t,y_t)\}$
\EndFor\\
\Return{$\{x_t\}_{t=1}^T$
} 
\end{algorithmic}
\end{algorithm}

\noindent\textbf{Output–Coverage Acquisition.}
We define the optimistic feasibility indicator in the objective space
\[
\tilde Z_{t-1}(x) \;=\;
\begin{cases}
1, & \text{if } U^{(i)}_{t-1}(x)\ge \tau_i \ \ \forall i,\\[2pt]
0, & \text{otherwise.}
\end{cases}
\]
Next, we define the incremental coverage gain at resolution $r$ which is our acquisition function used in step 3 in Algorithm~\ref{alg:moccas}:
\begin{align}
\alpha^{\MOCCAS}_{t-1}(x)
&:= \tilde Z_{t-1}(x)\cdot
\Vol \Bigl(\bigl(\Ball{r}{U_{t-1}(x)}\cap\cS\bigr)\setminus \Ball{r}{Z_{t-1}}\Bigr),\label{eq:hard-acq}
\end{align}
where $\Ball{r}{Z_{t-1}} := \bigcup_{s=1}^{t-1}\Ball{r}{y_s}$.
Value is gained only if $x$ is optimistically feasible, and then only by the new $r$-ball volume supplied within $\cS$.
We prefer the sample whose predicted outcome $U_{t-1}(x)$ is farthest (in $\ell_2$) from the existing set $Z_{t-1}$ in order to diversify among ties.

\noindent\textbf{Smoothing for efficient optimization.}
We substitute smooth surrogates (e.g., soft margins with respect to differentiable $r$-ball kernels and thresholds $\tau$) for the hard feasibility test and set-difference to enable fast and gradient-based maximization of the acquisition function.
This provides a smooth surrogate approximation of the discrete gain and could be optimized much more efficiently with the help of known gradient-based optimization methods like L-BFGS. The formal construction of smoothing is presented in Section \ref{sec:theory}.

\section{Efficient Optimization for Practice}\label{sec:theory}

In this section, we analyze and efficiently optimize Algorithm~\ref{alg:moccas}. Here, we delineate both the exact (hard) acquisition with an optimistic feasibility guard and a fully smooth surrogate function utilized for inner optimization. More analyses of acquisition approximation error and time complexity of \textsc{MOC-CAS} are deferred to the Appendix.

Beginning from the hard geometric definition (eq. \eqref{eq:hard-acq}), Step~A in Subsection~\ref{sec:theory_hard} converts the set difference into a local Gaussian average. Then, Step~B in Subsection~\ref{sec:theory_hard} replaces the hard feasibility by a smooth multi-threshold gate evaluated at the local center. Finally, Step~C in Subsection~\ref{sec:theory_hard} replaces the union by a soft sum whose local average yields a closed-form expression. The resulting smooth acquisition function (eq. \eqref{eq:soft-acq}) is cheap to evaluate and also easy to differentiate, thus yielding an efficient inner maximization routine for \MOCCAS{}. 

\subsection{From Hard to Soft Indicators: Deriving the MOC-CAS Acquisition}\label{sec:theory_hard}

The hard new-coverage objective in output space at iteration $t$ is given in eq. \eqref{eq:hard-acq}, that is, the $\cS$-constrained volume newly covered by the $r$-ball around the optimistic prediction $U_{t-1}(x)$ while discarding the region already covered by the previous outputs $\{y_s\}_{s=1}^{t-1}$.

\paragraph{Step A: Replace the Ball by a Mass-preserved Normalized Gaussian Kernel}
We write the set-volume in eq. \eqref{eq:hard-acq} as an integral over three hard indicators:
\begin{multline}\label{eq:int-indicators}
\alpha^{\MOCCAS}_{t-1}(x)
= \tilde Z_{t-1}(x) 
\int_{\R^m}
\inds{\Ball{r}{U_{t-1}(x)}}{z}\,
\inds{\cS}{z}\,\\[-2pt]
\cdot\Bigl(1-\inds{\Ball{r}{Z_{t-1}}}{z}\Bigr)\,dz.
\end{multline}

The three factors inside the integral in \eqref{eq:int-indicators} correspond to the new $r$-ball, the satisfactory set, and the not-yet-covered region, respectively.

We introduce two unit–mass radial kernels in output space. One is the uniform kernel on the open ball
\begin{align*}
    u_r(z-U) :=\frac{1}{V_m(r)}\,\ind{\norm{z-U}<r},\qquad V_m(r) :=\frac{\pi^{m/2}}{\Gamma \bigl(\tfrac{m}{2}+1\bigr)} r^m,
\end{align*}
where $\Gamma$ is the Gamma function, and another one is the Gaussian kernel
\[
\kappa_r(z-U)
:=\Nc \bigl(z;\,U,\,r^2 I\bigr),
\qquad \int \kappa_r(z-U)\,dz=1.
\]
We then make the standard mass-preserving substitution
\begin{equation*}\label{eq:mass-preserving}
\inds{\Ball{r}{U}}{z}\ \approx\ V_m(r)\,\kappa_r(z-U),
\end{equation*}
which yields
\begin{equation}\label{eq:local-expectation}
\begin{aligned}
& \quad \alpha^{\MOCCAS}_{t-1}(x)\\
&\approx \tilde Z_{t-1}(x)\,V_m(r) \int_{\R^m} \kappa_r(z-U)\,\inds{\cS}{z}\,
\bigl(1-\inds{\Ball{r}{Z_{t-1}}}{z}\bigr)\,dz\\
&= \tilde Z_{t-1}(x)\,V_m(r)\,
\E_{Z\sim\Nc(U,r^2 I)} \Big[\inds{\cS}{Z}\bigl(1-\inds{\Ball{r}{Z_{t-1}}}{Z}\bigr)\Big],
\end{aligned}
\end{equation}
where $U:=U_{t-1}(x)$. Therefore, Step~A converts the hard geometry into a local Gaussian average around $U$.

\paragraph{Step B: Soften the Satisfactory Gate}
We replace the hard indicator of the thresholded orthant $\inds{\cS}{z}=\prod_{i=1}^m \ind{z_i\ge \tau_i}$ by a smooth \emph{probit gate} with softness $\lambda>0$:
\begin{equation}\label{eq:psat}
p_{\mathrm{sat}}(z)
\ :=\
\prod_{i=1}^m \Phi \Big(\frac{z_i-\tau_i}{\lambda}\Big),
\end{equation}
where $\Phi$ is the standard normal cumulative distribution function (CDF). Using the narrow and centered Gaussian kernel from eq. \eqref{eq:local-expectation}, we approximate the local average of the satisfactory gate by its value at the center:
\begin{equation}\label{eq:local-constancy}
\E_{Z\sim\Nc(U,r^2 I)} \big[p_{\mathrm{sat}}(Z)\big]\ \approx\ p_{\mathrm{sat}}(U).
\end{equation}
Substituting eq. \eqref{eq:psat} and \eqref{eq:local-constancy} into eq. \eqref{eq:local-expectation} gives
\begin{align*}
& \quad \alpha^{\MOCCAS}_{t-1}(x)\\
& \approx \tilde Z_{t-1}(x)\,V_m(r)\,p_{\mathrm{sat}}(U)\;
\E_{Z\sim\Nc(U,r^2 I)}[1-\inds{\Ball{r}{Z_{t-1}}}{Z}].
\end{align*}

\paragraph{Step C: Soften the Set Difference (Union) and Derive a Closed-Form Expectation}
$\inds{\Ball{r}{Z_{t-1}}}{z}=\ind{z\in\cup_{s=1}^{t-1}\Ball{r}{y_s}}$ is the exact membership of the already-covered region.
We utilize a bounded soft union by summing the normalized Gaussian kernels centered at the previous outputs, which yields the “soft-OR” of the covered neighborhoods:
\begin{equation*}\label{eq:soft-union}
\inds{\Ball{r}{Z_{t-1}}}{z} \approx \sum_{s=1}^{t-1}\kappa_r(z-y_s).
\end{equation*}
Under the local Gaussian in eq. \eqref{eq:local-expectation}, the expectation of each term is reduced to a product–integral of two Gaussian distributions:
\[
\begin{aligned}
\E_{Z\sim\Nc(U,r^2 I)} \big[\kappa_r(Z-y_s)\big]
&= \int \Nc(z;U,r^2 I)\,\Nc(z;y_s,r^2 I)\,dz\\
&= \Nc \big(U;\,y_s,\,2r^2 I\big)\\
&= c_m(r)\,\exp \Bigl(-\tfrac{\norm{U-y_s}^2}{4r^2}\Bigr).
\end{aligned}
\]
where $c_m(r):=(4\pi r^2)^{-m/2}$. Therefore,
\begin{equation}\label{eq:novelty}
\begin{aligned}
\E_{Z\sim\Nc(U,r^2 I)} \Big[\,1-\inds{\Ball{r}{Z_{t-1}}}{Z}\,\Big]
&\approx 1-\sum_{s=1}^{t-1} c_m(r)\,
\exp \Bigl(-\tfrac{\norm{U-y_s}^2}{4r^2}\Bigr)\\
&=: n(U).
\end{aligned}
\end{equation}

\subsection{Soft MOC-CAS Acquisition and Efficient Maximization}
Once we combine eq. \eqref{eq:local-expectation}, \eqref{eq:psat}, and \eqref{eq:novelty}, it produces the fully softened acquisition function:
\begin{equation}\label{eq:soft-acq}
\widetilde{\alpha}^{\MOCCAS}_{t-1}(x)
\ :=\
V_m(r)\;\underbrace{p_{\mathrm{sat}} \big(U_{t-1}(x)\big)}_{\text{soft feasibility}}\;
\underbrace{n \big(U_{t-1}(x)\big)}_{\text{soft new-coverage}}.
\end{equation}

This expression is not expensive to evaluate. This is because $V_m(r)$ is a constant given $r$ and $m$. Also, $p_{\mathrm{sat}}(U)$ is a product of 1-dimensional CDFs at $U$. Furthermore, $n(U)$ is a short sum of exponentials in squared distances to the previous outputs.
The hard gate $\tilde Z_{t-1}(x)$ appears in the exact (hard) acquisition function defined in eq. \eqref{eq:hard-acq} to enforce optimistic feasibility.
For efficient and fully differentiable optimization, we drop this binary factor in eq. \eqref{eq:soft-acq} and utilize the smooth feasibility term $p_{\mathrm{sat}}(U)$.
Empirically, both variants behave identically. By default, we utilize \eqref{eq:soft-acq} for inner optimization.

\paragraph{Gradients for inner optimization.}
Let $U=U_{t-1}(x)$ and we write $V_m(r)$ once as a constant factor.
\[
\nabla_U p_{\mathrm{sat}}(U)
\;=\;
p_{\mathrm{sat}}(U)\,\Bigl[\,
\frac{\phi\bigl(\tfrac{U_1-\tau_1}{\lambda}\bigr)}{\lambda\,\Phi\bigl(\tfrac{U_1-\tau_1}{\lambda}\bigr)},
\ldots,
\frac{\phi\bigl(\tfrac{U_m-\tau_m}{\lambda}\bigr)}{\lambda\,\Phi\bigl(\tfrac{U_m-\tau_m}{\lambda}\bigr)}
\Bigr]^\top,
\]

Here, we use the standard normal CDF $\Phi(u)$ and PDF $\phi(u)$
\[
\Phi(u)=\int_{-\infty}^{u}\frac{1}{\sqrt{2\pi}}e^{-t^{2}/2}\,dt,
\qquad
\phi(u)=\frac{1}{\sqrt{2\pi}}e^{-u^{2}/2},
\]
and in the gradient, the argument is \(u_i=\tfrac{U_i-\tau_i}{\lambda}\).

\[
\nabla_U n(U)
\;=\;
\sum_{s=1}^{t-1} c_m(r)\,\exp \Big(-\tfrac{\norm{U-y_s}^2}{4r^2}\Big)\;\frac{U-y_s}{2r^2}.
\]
By the chain rule, with $U_i(x)=\mu^{(i)}_{t-1}(x)+\sqrt{\beta_t}\,\sigma^{(i)}_{t-1}(x)$,
\begin{equation}\label{eq:chain-rule}
\begin{aligned}
\nabla_x \widetilde{\alpha}^{\MOCCAS}_{t-1}(x)
&= V_m(r)\,\Big(\nabla_U p_{\mathrm{sat}}(U)\,n(U)
+ p_{\mathrm{sat}}(U)\,\nabla_U n(U)\Big)^\top\\
&\hspace{3.2em}\cdot\,\nabla_x U(x).
\end{aligned}
\end{equation}
where, for each $i$,
\[
\nabla_x U_i(x)
=\nabla_x \mu^{(i)}_{t-1}(x)+\sqrt{\beta_t}\,\nabla_x \sigma^{(i)}_{t-1}(x).
\]
These derivatives are already available in standard GP toolkits. Eq.~\eqref{eq:soft-acq} together with eq. \eqref{eq:chain-rule} yields a smooth objective and explicit gradients for multi-start L-BFGS (or similar methods) over $x\in\X$. The best $x_t$ is then chosen by $x_t= \arg\max_{x\in\X}\ \widetilde{\alpha}^{\MOCCAS}_{t-1}(x)$.

\section{Experiments}\label{sec:exp}

\subsection{Experimental Settings}

\begin{figure*}[t]
    \centering
    \includegraphics[width=0.9\textwidth]{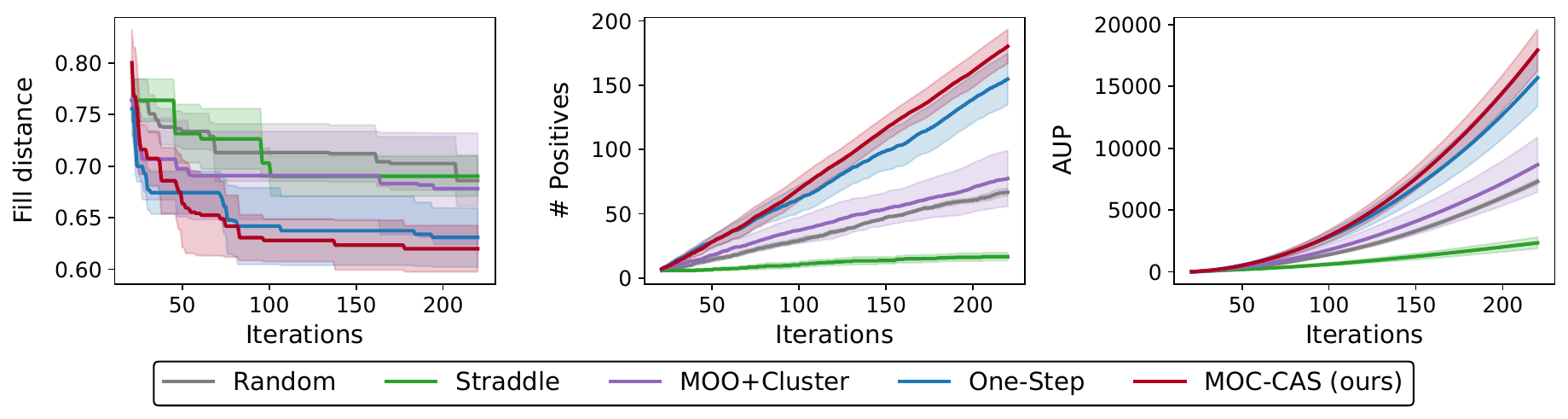}
    \caption{Quantitative comparison on the \textit{SARS-CoV-2 3CLPro} target dataset.}
    \label{fig:covid}
    \Description{Quantitative comparison on the SARS-CoV-2 3CLPro target dataset using three metrics over iterations. The figure contains multiple line plots with a legend for different methods. Panels report fill distance, cumulative number of feasible discoveries, and the area under the positives curve. Lower fill distance and higher discovery and area-under-curve indicate better performance.}
\end{figure*}

\begin{figure*}[!htbp]
    \centering
    \includegraphics[width=0.9\textwidth]{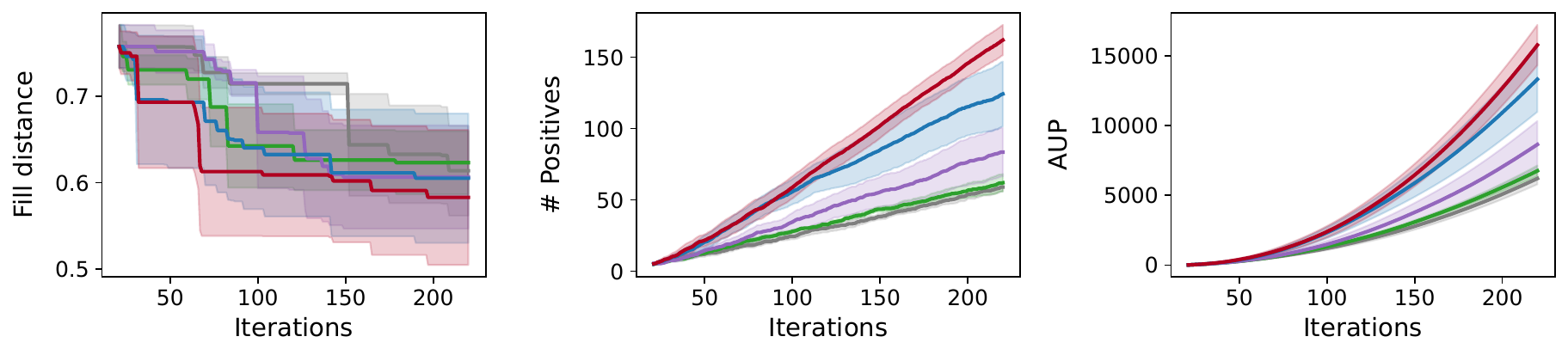}
    \includegraphics[width=0.9\textwidth]{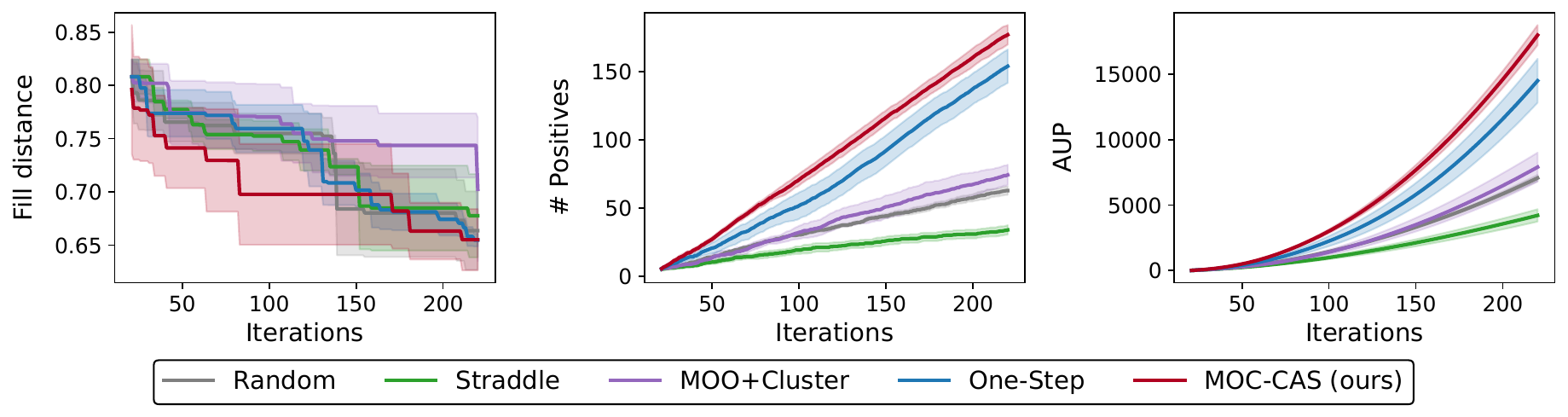}
    \caption{Quantitative comparison on the \textit{Cancer RTCB} (top row) and \textit{Cancer 6T2W} (bottom row) target datasets.}
    \label{fig:cancer_6t2w}
    \Description{Two-row grid of quantitative comparisons on cancer target datasets. The top row corresponds to the RTCB target and the bottom row corresponds to the 6T2W target. For each target, multiple methods are compared with line plots over iterations using fill distance, cumulative number of feasible discoveries, and the area under the positives curve.}
\end{figure*}

\noindent \textbf{Datasets.} 
We use the DrugImprover dataset~\citep{liu2023drugimprover,liu2025drugimprovergpt} that comprises curated, orderable subsets of the ZINC15 compound library, where 1 million in-stock or quickly shippable molecules were selected per target for structure-based drug discovery. Protein pockets were defined using crystallographic ligand data or FPocket, and a single binding site was used for docking per protein. The dataset includes docking scores (via OpenEye FRED) and trained surrogate models (with validation $R^2$ of 0.842 for 3CLPro and 0.73 for RTCB) for 24 SARS-CoV-2 protein (e.g., \textit{3CLPro}, \textit{PLPro}, \textit{RDRP}, \textit{NSPs}) and 5 human cancer-related targets (e.g., \textit{6T2W}, \textit{RTCB}, \textit{WRN}). Provided files include Simplified Molecular Input Line Entry System (SMILES)-docking score pairs, receptor structures, model weights, and inference code, supporting applications in high-throughput virtual screening and predictive modeling.
From the available targets, we use three cancer proteins \textit{6T2W}, \textit{RTCB}, and \textit{WRN}, and one SARS-CoV-2 protease target \textit{3CLPro}. 

\noindent \textbf{Baselines.}
We benchmark \MOCCAS{} against the following four baseline methods on all the four drug–discovery tasks to delineate the comparison. 

 \textsc{Random.} We utilize the uniform random selection method from the remaining pool each round. 

 \textsc{One-Step.} We apply the one–step active search method that selects the sample maximizing the predicted feasibility probability \(p(Z{=}1\mid\cD_{t-1})\), calculated as the product of per–objective Gaussian tail probabilities under independent GPs \citep{garnett2012bayesian}. 

 \textsc{Straddle.} We implement a multiobjective extension of STRADDLE \citep{bryan2005active} for level–set estimation that alternates the targeted objective \(i\in[m]\) each iteration and scores candidates by $s_i(x)=\kappa \sigma^{(i)}_{t-1}(x)- |\mu^{(i)}_{t-1}(x)-\tau_i |, \kappa = 1.96$,
then selects the maximizer using the shared shortlist. 

 \textsc{MOO+Cluster.} We implement the multiobjective BO with clustering that develops optimistic outcome predictions \(U_{t-1}(x)\) using UCB, keeps optimistically feasible candidates, clusters their estimated outcomes with \(k\)-means, scores each cluster by its uncovered feasible mass within radius \(r\) (via neighbor counts in objective space), then selects within the best cluster the point farthest (in objective space) from the previously observed outcomes. 

All baseline methods share the same GP modeling pipeline, budget, initializations, and shortlist settings as \MOCCAS{} for ensuring controlled and fair comparisons.

\noindent \textbf{Evaluation.}
As discussed in Section~\ref{sec:pre:problem}, we report the number of positives (samples with all objectives \(\ge\tau\)), AUP, and fill distance inside the feasible outcome region, over 220 iterations. We finally aggregate the results as mean\(\pm\)adjusted standard error over four different trials for each experiment (i.e., $\mathrm{mean} \pm \mathrm{std}/\sqrt{4}$). 

\subsection{Results on \textit{SARS-CoV-2} Dataset}

We test all five methods on the \textit{SARS-CoV-2 3CLPro} target dataset. \MOCCAS{} achieves the best results, as evidenced in Figure~\ref{fig:covid}, where it consistently achieves the lowest fill distance and highest AUP and number of positives across multiple trials, outperforming all other algorithms. \textsc{One-step} active search method is competitive but underperforms compared to our \MOCCAS{} method. Overall, \textsc{straddle}, \textsc{random search}, and \textsc{MOO+Cluster} did not perform well in terms of the three evaluation criteria where \textsc{straddle} performs worst among all methods in terms of number of positives and AUP.

\textsc{One-step} selects points that maximize the estimated feasibility probability, therefore it tends to exploit the high-feasibility region (where feasibility probability is highest) early. This makes it competitive on number of positives, however, it can concentrate evaluations in a smaller portion of the feasible objective space, which is consistent with its slightly worse fill-distance trajectory than \MOCCAS{}. In contrast, \MOCCAS{} explicitly rewards coverage gain in the feasible objective space, so once it finds a feasible neighborhood, it is driven to expand into new feasible neighborhoods rather than repeatedly sampling near already-covered outcomes. This is exactly the behavior we want in molecular screening: beyond finding many feasible candidates, we want them to be diverse in objective space to support down-selection and follow-up synthesis. Finally, the weaker baselines align with their design intent: \textsc{straddle} is primarily a level-set (boundary) learner, therefore it focuses on uncertain points near thresholds and consequently sacrifices feasible yield and AUP. \textsc{MOO+Cluster} relies on clustering of optimistic predictions and is less effective when posterior uncertainty is high, whereas \textsc{random search} provides a lower-bound reference.   

\subsection{Results on \textit{Cancer} Dataset}

We test all five methods on three cancer proteins \textit{6T2W}, \textit{RTCB}, and \textit{WRN} target datasets. Again, our \MOCCAS{} turns out to be the best method in terms of all the evaluation criteria, as evidenced in Figure~\ref{fig:cancer_6t2w}. The performance gap between \MOCCAS{} and the baselines turns out to be statistically significant on the \textit{6T2W} dataset in Figure~\ref{fig:cancer_6t2w} in terms of AUP and number of positives - indicating that feasibility-only exploitation is not sufficient. The most plausible explanation is geometric: this target behaves like a harder coverage problem where feasible outcomes are either sparser or more fragmented in objective space, so repeatedly sampling the most confident feasible area leaves other feasible neighborhoods undiscovered. \MOCCAS{} is explicitly designed to exploit information from the posterior GPs and the history of discovered feasible outcomes, so its advantage grows as the posteriors sharpen and the soft feasibility/coverage terms become more accurate. In the initial iterations, all methods operate in a highly uncertain regime with very sparse data, so even sophisticated acquisitions behave close to exploratory baselines, which explains the small performance gaps.

Results on \textit{WRN} dataset are deferred to the Appendix which show a similar pattern. Among the baseline methods, \textsc{One-step} active search performed better than the other baselines across all the Cancer datasets. 

\begin{figure*}[t]
    \centering
    \includegraphics[width=0.9\textwidth]{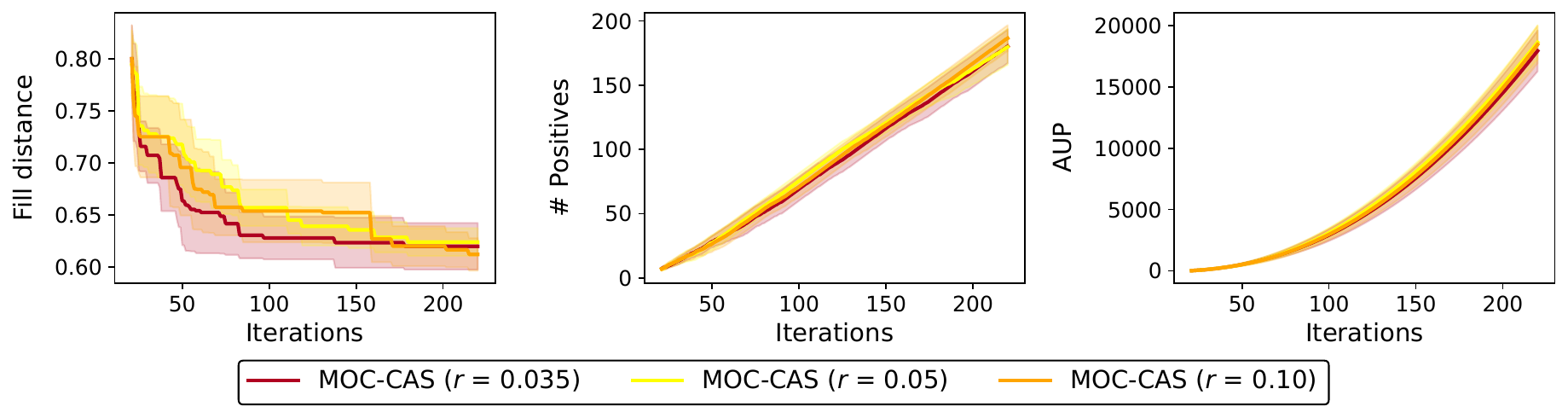}
    \caption{Ablation of coverage resolution $r$ on the \textit{SARS-CoV-2 3CLpro} dataset.}
    \label{fig:abl_r}
    \Description{Ablation study of the coverage resolution parameter r on the SARS-CoV-2 3CLpro dataset. The figure compares several r settings using line plots over iterations for three metrics: fill distance, cumulative number of feasible discoveries, and the area under the positives curve.}
\end{figure*}

\begin{figure*}[!htbp]
    \centering
    \includegraphics[width=0.9\textwidth]{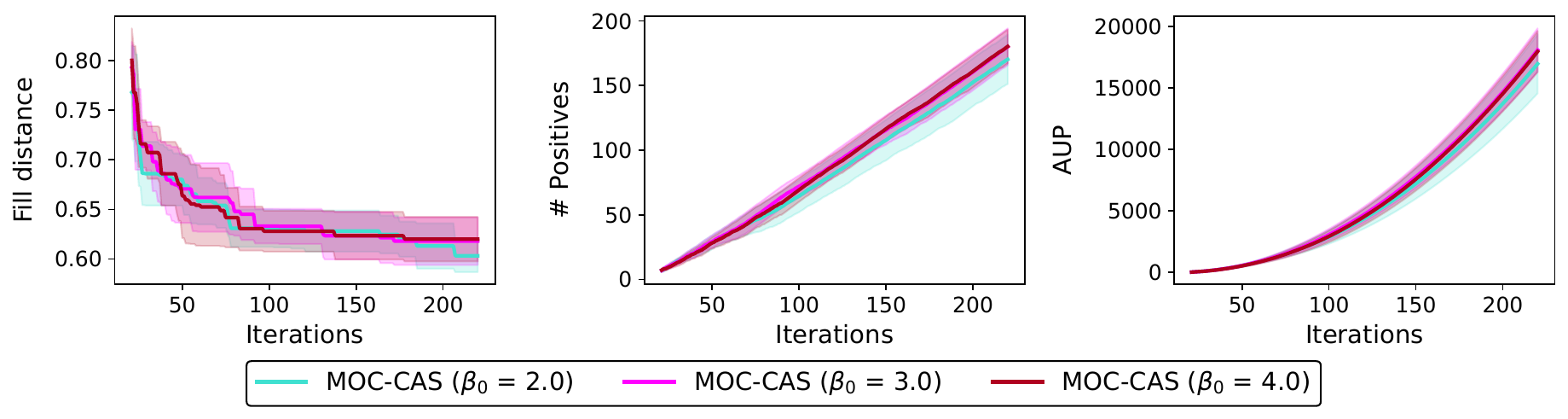}
    \caption{Ablation of optimism schedule $\beta_t$ on the \textit{SARS-CoV-2 3CLpro} dataset.}
    \label{fig:abl_beta}
    \Description{Ablation study of the optimism schedule on the SARS-CoV-2 3CLpro dataset. Multiple optimism schedules are compared with line plots over iterations for fill distance, cumulative number of feasible discoveries, and the area under the positives curve.}
\end{figure*}

\subsection{Ablation Studies}

\MOCCAS{} has two important hyperparameters carrying clear geometric meaning which govern its empirical efficiency and behavior. The coverage resolution, $r$ assigns the granularity at which objective–space diversity is rewarded. Therefore, changes in the values of chosen $r$ will have influence on the quality of dispersion, covered volume and cumulative positive samples found. Also, the optimism schedule, $\beta_t$ governs how strongly the UCB gate considers feasibility, trading exploration close to the thresholds against conservative consolidation and so its time profile forms early vs. late discovery and therefore AUP. Since objective scales, threshold calibrations, and noise levels vary across the targets (e.g., cancer vs.\ viral proteins), a fixed setting is not always universally optimal. Therefore, we ablate $r$ and $\beta_t$ for characterizing sensitivity by reporting influences on all the evaluation criteria.

\noindent \textbf{Coverage resolution ($r$).}
We study the effect of the coverage resolution $r$ by fixing all hyperparameters (including the UCB schedule) and reusing the same seeds and initial indices across different trials, then sweeping $r\in\{0.035,\,0.05,\,0.10\}$ on the \textit{SARS-CoV-2 3CLpro} dataset. We show the results in Figure \ref{fig:abl_r}. Across our runs, the highest resolution ($r$ = 0.10) delivered the most favorable trade–off yielding slightly lower fill distance and higher number of positives than others while maintaining competitive AUP, whereas the intermediate resolution ($r$ = 0.05) delivered slightly higher AUP. Figure \ref{fig:abl_r} also shows that the smaller radius ($r$ = 0.035) comparatively under–covers the feasible region.

\noindent \textbf{Optimism schedule ($\beta_t$).}
In our implementation, the time-varying confidence is $\beta_t = \beta_0 \cdot a_t$,
where the annealing multiplier $a_t \in (0,1]$ is held fixed across runs. The ablation changes only $\beta_0$ while keeping $a_t$ constant. To investigate exploration--exploitation, we fix the coverage radius and all other settings and then sweep the initial optimism level $\beta_0\in\{2.0,\,3.0,\,4.0\}$ with the same annealing schedule (with constant seeds and initial indices across four different trials). 
Figure~\ref{fig:abl_beta} investigates the optimism schedule: a larger $\beta_0$ early encourages broader exploration that can escape low-quality pockets, whereas smaller values promote later consolidation within the feasible region. Empirically, Figure~\ref{fig:abl_beta} shows that increasing $\beta_0$ modestly improves AUP and cumulative positives, while smaller $\beta_0$ tends to slightly reduce fill distance.

\section{Conclusion}\label{sec:con}

We address a practically important setting common in drug discovery and materials design: when success depends on meeting multiple performance thresholds, the objective is not to trace a Pareto front or spread samples in input space, but to rapidly identify a small, representative set whose outcomes broadly cover the feasible region. Such sets enable faster downstream evaluation and decision-making, accelerating scientific iteration under tight experimental budgets.

We formulate this as a multi-objective coverage problem and propose \MOCCAS, an acquisition strategy that integrates Gaussian process UCB predictions with an incremental coverage objective. To enable efficient inner optimization, we introduce a smoothed relaxation of the hard geometric objective, yielding a differentiable surrogate with closed-form components and simple gradients. Across large-scale protein–target benchmarks for cancer and SARS–CoV–2, \MOCCAS consistently achieves better coverage–quality tradeoffs than strong baselines, improving early discovery (higher cumulative positives and AUP) while maintaining diversity (lower fill distance). Our study employs independent Gaussian processes for each objective, though richer outcome models, such as deep kernels or correlated multi-output priors, represent promising directions for future work.

\begin{acks}
This work was partially supported by the IBM-UAlbany CEAIS Seed Grant 1201104-1-102522. The authors would like to thank anonymous reviewers and the area chair for helpful comments that improve this paper.

\end{acks}

\bibliographystyle{ACM-Reference-Format} 
\bibliography{ref}


\begin{thebibliography}{34}


\ifx \showCODEN    \undefined \def \showCODEN     #1{\unskip}     \fi
\ifx \showDOI      \undefined \def \showDOI       #1{#1}\fi
\ifx \showISBNx    \undefined \def \showISBNx     #1{\unskip}     \fi
\ifx \showISBNxiii \undefined \def \showISBNxiii  #1{\unskip}     \fi
\ifx \showISSN     \undefined \def \showISSN      #1{\unskip}     \fi
\ifx \showLCCN     \undefined \def \showLCCN      #1{\unskip}     \fi
\ifx \shownote     \undefined \def \shownote      #1{#1}          \fi
\ifx \showarticletitle \undefined \def \showarticletitle #1{#1}   \fi
\ifx \showURL      \undefined \def \showURL       {\relax}        \fi
\providecommand\bibfield[2]{#2}
\providecommand\bibinfo[2]{#2}
\providecommand\natexlab[1]{#1}
\providecommand\showeprint[2][]{arXiv:#2}

\bibitem[\protect\citeauthoryear{Bogunovic, Scarlett, Krause, and Cevher}{Bogunovic et~al\mbox{.}}{2016}]%
        {bogunovic2016truncated}
\bibfield{author}{\bibinfo{person}{Ilija Bogunovic}, \bibinfo{person}{Jonathan Scarlett}, \bibinfo{person}{Andreas Krause}, {and} \bibinfo{person}{Volkan Cevher}.} \bibinfo{year}{2016}\natexlab{}.
\newblock \showarticletitle{Truncated variance reduction: A unified approach to bayesian optimization and level-set estimation}.
\newblock \bibinfo{journal}{\emph{Advances in neural information processing systems}}  \bibinfo{volume}{29} (\bibinfo{year}{2016}).
\newblock


\bibitem[\protect\citeauthoryear{Bryan, Nichol, Genovese, Schneider, Miller, and Wasserman}{Bryan et~al\mbox{.}}{2005}]%
        {bryan2005active}
\bibfield{author}{\bibinfo{person}{Brent Bryan}, \bibinfo{person}{Robert~C Nichol}, \bibinfo{person}{Christopher~R Genovese}, \bibinfo{person}{Jeff Schneider}, \bibinfo{person}{Christopher~J Miller}, {and} \bibinfo{person}{Larry Wasserman}.} \bibinfo{year}{2005}\natexlab{}.
\newblock \showarticletitle{Active learning for identifying function threshold boundaries}.
\newblock \bibinfo{journal}{\emph{Advances in neural information processing systems}}  \bibinfo{volume}{18} (\bibinfo{year}{2005}).
\newblock


\bibitem[\protect\citeauthoryear{Daulton, Balandat, and Bakshy}{Daulton et~al\mbox{.}}{2021}]%
        {daulton2021parallel}
\bibfield{author}{\bibinfo{person}{Samuel Daulton}, \bibinfo{person}{Maximilian Balandat}, {and} \bibinfo{person}{Eytan Bakshy}.} \bibinfo{year}{2021}\natexlab{}.
\newblock \showarticletitle{Parallel bayesian optimization of multiple noisy objectives with expected hypervolume improvement}.
\newblock \bibinfo{journal}{\emph{Advances in neural information processing systems}}  \bibinfo{volume}{34} (\bibinfo{year}{2021}), \bibinfo{pages}{2187--2200}.
\newblock


\bibitem[\protect\citeauthoryear{Daulton, Eriksson, Balandat, and Bakshy}{Daulton et~al\mbox{.}}{2022}]%
        {daulton2022multi}
\bibfield{author}{\bibinfo{person}{Samuel Daulton}, \bibinfo{person}{David Eriksson}, \bibinfo{person}{Maximilian Balandat}, {and} \bibinfo{person}{Eytan Bakshy}.} \bibinfo{year}{2022}\natexlab{}.
\newblock \showarticletitle{Multi-objective bayesian optimization over high-dimensional search spaces}. In \bibinfo{booktitle}{\emph{Uncertainty in Artificial Intelligence}}. PMLR, \bibinfo{pages}{507--517}.
\newblock


\bibitem[\protect\citeauthoryear{Emmerich, Deutz, and Klinkenberg}{Emmerich et~al\mbox{.}}{2011}]%
        {emmerich2011hypervolume}
\bibfield{author}{\bibinfo{person}{Michael~TM Emmerich}, \bibinfo{person}{Andr{\'e}~H Deutz}, {and} \bibinfo{person}{Jan~Willem Klinkenberg}.} \bibinfo{year}{2011}\natexlab{}.
\newblock \showarticletitle{Hypervolume-based expected improvement: Monotonicity properties and exact computation}. In \bibinfo{booktitle}{\emph{2011 IEEE congress of evolutionary computation (CEC)}}. IEEE, \bibinfo{pages}{2147--2154}.
\newblock


\bibitem[\protect\citeauthoryear{Frazier}{Frazier}{2018}]%
        {frazier2018tutorial}
\bibfield{author}{\bibinfo{person}{Peter~I Frazier}.} \bibinfo{year}{2018}\natexlab{}.
\newblock \showarticletitle{A tutorial on Bayesian optimization}.
\newblock \bibinfo{journal}{\emph{arXiv preprint arXiv:1807.02811}} (\bibinfo{year}{2018}).
\newblock


\bibitem[\protect\citeauthoryear{Garnett, Krishnamurthy, Xiong, Schneider, and Mann}{Garnett et~al\mbox{.}}{2012}]%
        {garnett2012bayesian}
\bibfield{author}{\bibinfo{person}{Roman Garnett}, \bibinfo{person}{Yamuna Krishnamurthy}, \bibinfo{person}{Xuehan Xiong}, \bibinfo{person}{Jeff Schneider}, {and} \bibinfo{person}{Richard Mann}.} \bibinfo{year}{2012}\natexlab{}.
\newblock \showarticletitle{Bayesian optimal active search and surveying}.
\newblock \bibinfo{journal}{\emph{arXiv preprint arXiv:1206.6406}} (\bibinfo{year}{2012}).
\newblock


\bibitem[\protect\citeauthoryear{Gotovos, Casati, Hitz, and Krause}{Gotovos et~al\mbox{.}}{2013}]%
        {gotovos2013active}
\bibfield{author}{\bibinfo{person}{Alkis Gotovos}, \bibinfo{person}{Nathalie Casati}, \bibinfo{person}{Gregory Hitz}, {and} \bibinfo{person}{Andreas Krause}.} \bibinfo{year}{2013}\natexlab{}.
\newblock \showarticletitle{Active learning for level set estimation}. In \bibinfo{booktitle}{\emph{International Joint Conference on Artificial Intelligence}}. \bibinfo{pages}{1344--1350}.
\newblock


\bibitem[\protect\citeauthoryear{Guyon, Cawley, Dror, and Lemaire}{Guyon et~al\mbox{.}}{2011}]%
        {guyon2011results}
\bibfield{author}{\bibinfo{person}{Isabelle Guyon}, \bibinfo{person}{Gavin~C Cawley}, \bibinfo{person}{Gideon Dror}, {and} \bibinfo{person}{Vincent Lemaire}.} \bibinfo{year}{2011}\natexlab{}.
\newblock \showarticletitle{Results of the active learning challenge}. In \bibinfo{booktitle}{\emph{Active Learning and Experimental Design workshop In conjunction with AISTATS 2010}}. JMLR Workshop and Conference Proceedings, \bibinfo{pages}{19--45}.
\newblock


\bibitem[\protect\citeauthoryear{Hern{\'a}ndez-Lobato, Hernandez-Lobato, Shah, and Adams}{Hern{\'a}ndez-Lobato et~al\mbox{.}}{2016}]%
        {hernandez2016predictive}
\bibfield{author}{\bibinfo{person}{Daniel Hern{\'a}ndez-Lobato}, \bibinfo{person}{Jose Hernandez-Lobato}, \bibinfo{person}{Amar Shah}, {and} \bibinfo{person}{Ryan Adams}.} \bibinfo{year}{2016}\natexlab{}.
\newblock \showarticletitle{Predictive entropy search for multi-objective bayesian optimization}. In \bibinfo{booktitle}{\emph{International conference on machine learning}}. PMLR, \bibinfo{pages}{1492--1501}.
\newblock


\bibitem[\protect\citeauthoryear{Iwazaki, Inatsu, and Takeuchi}{Iwazaki et~al\mbox{.}}{2020}]%
        {iwazaki2020bayesian}
\bibfield{author}{\bibinfo{person}{Shogo Iwazaki}, \bibinfo{person}{Yu Inatsu}, {and} \bibinfo{person}{Ichiro Takeuchi}.} \bibinfo{year}{2020}\natexlab{}.
\newblock \showarticletitle{Bayesian experimental design for finding reliable level set under input uncertainty}.
\newblock \bibinfo{journal}{\emph{IEEE Access}}  \bibinfo{volume}{8} (\bibinfo{year}{2020}), \bibinfo{pages}{203982--203993}.
\newblock


\bibitem[\protect\citeauthoryear{Knowles}{Knowles}{2006}]%
        {knowles2006parego}
\bibfield{author}{\bibinfo{person}{Joshua Knowles}.} \bibinfo{year}{2006}\natexlab{}.
\newblock \showarticletitle{ParEGO: A hybrid algorithm with on-line landscape approximation for expensive multiobjective optimization problems}.
\newblock \bibinfo{journal}{\emph{IEEE transactions on evolutionary computation}} \bibinfo{volume}{10}, \bibinfo{number}{1} (\bibinfo{year}{2006}), \bibinfo{pages}{50--66}.
\newblock


\bibitem[\protect\citeauthoryear{Komiyama, Malkomes, Cheng, and McCourt}{Komiyama et~al\mbox{.}}{2022}]%
        {komiyama2022bridging}
\bibfield{author}{\bibinfo{person}{Junpei Komiyama}, \bibinfo{person}{Gustavo Malkomes}, \bibinfo{person}{Bolong Cheng}, {and} \bibinfo{person}{Michael McCourt}.} \bibinfo{year}{2022}\natexlab{}.
\newblock \showarticletitle{Bridging offline and online experimentation: Constraint active search for deployed performance optimization}.
\newblock \bibinfo{journal}{\emph{Transactions on Machine Learning Research}} (\bibinfo{year}{2022}).
\newblock


\bibitem[\protect\citeauthoryear{Landrum and contributors}{Landrum and contributors}{[n.d.]}]%
        {rdkit}
\bibfield{author}{\bibinfo{person}{Greg Landrum} {and} \bibinfo{person}{contributors}.} \bibinfo{year}{[n.d.]}\natexlab{}.
\newblock \bibinfo{title}{{RDKit}: Open-source cheminformatics}.
\newblock \bibinfo{howpublished}{\url{https://www.rdkit.org}}.
\newblock
\newblock
\shownote{Accessed: 2025-10-08.}


\bibitem[\protect\citeauthoryear{Laumanns and Ocenasek}{Laumanns and Ocenasek}{2002}]%
        {laumanns2002bayesian}
\bibfield{author}{\bibinfo{person}{Marco Laumanns} {and} \bibinfo{person}{Jiri Ocenasek}.} \bibinfo{year}{2002}\natexlab{}.
\newblock \showarticletitle{Bayesian optimization algorithms for multi-objective optimization}. In \bibinfo{booktitle}{\emph{International Conference on Parallel Problem Solving from Nature}}. Springer, \bibinfo{pages}{298--307}.
\newblock


\bibitem[\protect\citeauthoryear{Li, Cho, and Liu}{Li et~al\mbox{.}}{2025a}]%
        {li2025none}
\bibfield{author}{\bibinfo{person}{Diantong Li}, \bibinfo{person}{Kyunghyun Cho}, {and} \bibinfo{person}{Chong Liu}.} \bibinfo{year}{2025}\natexlab{a}.
\newblock \showarticletitle{None To Optima in Few Shots: Bayesian Optimization with MDP Priors}.
\newblock \bibinfo{journal}{\emph{arXiv preprint arXiv:2511.01006}} (\bibinfo{year}{2025}).
\newblock


\bibitem[\protect\citeauthoryear{Li, Zhang, Liu, and Chen}{Li et~al\mbox{.}}{2025b}]%
        {li2025constrained}
\bibfield{author}{\bibinfo{person}{Diantong Li}, \bibinfo{person}{Fengxue Zhang}, \bibinfo{person}{Chong Liu}, {and} \bibinfo{person}{Yuxin Chen}.} \bibinfo{year}{2025}\natexlab{b}.
\newblock \showarticletitle{Constrained Multi-objective Bayesian Optimization through Optimistic Constraints Estimation}. In \bibinfo{booktitle}{\emph{International Conference on Artificial Intelligence and Statistics}}. PMLR, \bibinfo{pages}{370--378}.
\newblock


\bibitem[\protect\citeauthoryear{Liu, Jiang, Chen, Yang, Chen, Foster, and Stevens}{Liu et~al\mbox{.}}{2025}]%
        {liu2025drugimprovergpt}
\bibfield{author}{\bibinfo{person}{Xuefeng Liu}, \bibinfo{person}{Songhao Jiang}, \bibinfo{person}{Siyu Chen}, \bibinfo{person}{Zhuoran Yang}, \bibinfo{person}{Yuxin Chen}, \bibinfo{person}{Ian Foster}, {and} \bibinfo{person}{Rick Stevens}.} \bibinfo{year}{2025}\natexlab{}.
\newblock \showarticletitle{DrugImproverGPT: A Large Language Model for Drug Optimization with Fine-Tuning via Structured Policy Optimization}.
\newblock \bibinfo{journal}{\emph{arXiv preprint arXiv:2502.07237}} (\bibinfo{year}{2025}).
\newblock


\bibitem[\protect\citeauthoryear{Liu, Jiang, Vasan, Brace, Gokdemir, Brettin, Xia, Foster, and Stevens}{Liu et~al\mbox{.}}{2023}]%
        {liu2023drugimprover}
\bibfield{author}{\bibinfo{person}{Xuefeng Liu}, \bibinfo{person}{Songhao Jiang}, \bibinfo{person}{Archit Vasan}, \bibinfo{person}{Alexander Brace}, \bibinfo{person}{Ozan Gokdemir}, \bibinfo{person}{Thomas Brettin}, \bibinfo{person}{Fangfang Xia}, \bibinfo{person}{Ian Foster}, {and} \bibinfo{person}{Rick Stevens}.} \bibinfo{year}{2023}\natexlab{}.
\newblock \showarticletitle{DRUGIMPROVER: Utilizing reinforcement learning for multi-objective alignment in drug optimization}. In \bibinfo{booktitle}{\emph{NeurIPS 2023 Workshop on New Frontiers of AI for Drug Discovery and Development}}, Vol.~\bibinfo{volume}{1}. \bibinfo{pages}{8}.
\newblock


\bibitem[\protect\citeauthoryear{Malkomes, Cheng, Lee, and Mccourt}{Malkomes et~al\mbox{.}}{2021}]%
        {malkomes2021beyond}
\bibfield{author}{\bibinfo{person}{Gustavo Malkomes}, \bibinfo{person}{Bolong Cheng}, \bibinfo{person}{Eric~H Lee}, {and} \bibinfo{person}{Mike Mccourt}.} \bibinfo{year}{2021}\natexlab{}.
\newblock \showarticletitle{Beyond the pareto efficient frontier: Constraint active search for multiobjective experimental design}. In \bibinfo{booktitle}{\emph{International Conference on Machine Learning}}. PMLR, \bibinfo{pages}{7423--7434}.
\newblock


\bibitem[\protect\citeauthoryear{Mason, Jain, Mukherjee, Camilleri, Jamieson, and Nowak}{Mason et~al\mbox{.}}{2022}]%
        {mason2022nearly}
\bibfield{author}{\bibinfo{person}{Blake Mason}, \bibinfo{person}{Lalit Jain}, \bibinfo{person}{Subhojyoti Mukherjee}, \bibinfo{person}{Romain Camilleri}, \bibinfo{person}{Kevin Jamieson}, {and} \bibinfo{person}{Robert Nowak}.} \bibinfo{year}{2022}\natexlab{}.
\newblock \showarticletitle{Nearly Optimal Algorithms for Level Set Estimation}. In \bibinfo{booktitle}{\emph{International Conference on Artificial Intelligence and Statistics}}. PMLR, \bibinfo{pages}{7625--7658}.
\newblock


\bibitem[\protect\citeauthoryear{Misra, Roald, and Ng}{Misra et~al\mbox{.}}{2022}]%
        {misra2022learning}
\bibfield{author}{\bibinfo{person}{Sidhant Misra}, \bibinfo{person}{Line Roald}, {and} \bibinfo{person}{Yeesian Ng}.} \bibinfo{year}{2022}\natexlab{}.
\newblock \showarticletitle{Learning for constrained optimization: Identifying optimal active constraint sets}.
\newblock \bibinfo{journal}{\emph{INFORMS Journal on Computing}} \bibinfo{volume}{34}, \bibinfo{number}{1} (\bibinfo{year}{2022}), \bibinfo{pages}{463--480}.
\newblock


\bibitem[\protect\citeauthoryear{Paria, Kandasamy, and P{\'o}czos}{Paria et~al\mbox{.}}{2020}]%
        {paria2020flexible}
\bibfield{author}{\bibinfo{person}{Biswajit Paria}, \bibinfo{person}{Kirthevasan Kandasamy}, {and} \bibinfo{person}{Barnab{\'a}s P{\'o}czos}.} \bibinfo{year}{2020}\natexlab{}.
\newblock \showarticletitle{A flexible framework for multi-objective bayesian optimization using random scalarizations}. In \bibinfo{booktitle}{\emph{Uncertainty in Artificial Intelligence}}. PMLR, \bibinfo{pages}{766--776}.
\newblock


\bibitem[\protect\citeauthoryear{Park, Tagasovska, Maser, Ra, and Cho}{Park et~al\mbox{.}}{2023}]%
        {park2023botied}
\bibfield{author}{\bibinfo{person}{Ji~Won Park}, \bibinfo{person}{Nata{\v{s}}a Tagasovska}, \bibinfo{person}{Michael Maser}, \bibinfo{person}{Stephen Ra}, {and} \bibinfo{person}{Kyunghyun Cho}.} \bibinfo{year}{2023}\natexlab{}.
\newblock \showarticletitle{BOtied: Multi-objective Bayesian optimization with tied multivariate ranks}.
\newblock \bibinfo{journal}{\emph{arXiv preprint arXiv:2306.00344}} (\bibinfo{year}{2023}).
\newblock


\bibitem[\protect\citeauthoryear{Riniker and Landrum}{Riniker and Landrum}{2013}]%
        {riniker2013fingerprints}
\bibfield{author}{\bibinfo{person}{Sereina Riniker} {and} \bibinfo{person}{Gregory~A. Landrum}.} \bibinfo{year}{2013}\natexlab{}.
\newblock \showarticletitle{Open-source platform to benchmark fingerprints for ligand-based virtual screening}.
\newblock \bibinfo{journal}{\emph{Journal of Cheminformatics}} \bibinfo{volume}{5}, \bibinfo{number}{26} (\bibinfo{year}{2013}).
\newblock
\urldef\tempurl%
\url{https://doi.org/10.1186/1758-2946-5-26}
\showDOI{\tempurl}


\bibitem[\protect\citeauthoryear{Snoek, Larochelle, and Adams}{Snoek et~al\mbox{.}}{2012}]%
        {snoek2012practical}
\bibfield{author}{\bibinfo{person}{Jasper Snoek}, \bibinfo{person}{Hugo Larochelle}, {and} \bibinfo{person}{Ryan~P Adams}.} \bibinfo{year}{2012}\natexlab{}.
\newblock \showarticletitle{Practical bayesian optimization of machine learning algorithms}.
\newblock \bibinfo{journal}{\emph{Advances in neural information processing systems}}  \bibinfo{volume}{25} (\bibinfo{year}{2012}).
\newblock


\bibitem[\protect\citeauthoryear{Srinivas, Krause, Kakade, and Seeger}{Srinivas et~al\mbox{.}}{2010}]%
        {srinivas2010gaussian}
\bibfield{author}{\bibinfo{person}{Niranjan Srinivas}, \bibinfo{person}{Andreas Krause}, \bibinfo{person}{Sham Kakade}, {and} \bibinfo{person}{Matthias Seeger}.} \bibinfo{year}{2010}\natexlab{}.
\newblock \showarticletitle{Gaussian process optimization in the bandit setting: no regret and experimental design}. In \bibinfo{booktitle}{\emph{International Conference on International Conference on Machine Learning}}. \bibinfo{pages}{1015--1022}.
\newblock


\bibitem[\protect\citeauthoryear{Sui, Gotovos, Burdick, and Krause}{Sui et~al\mbox{.}}{2015}]%
        {sui2015safe}
\bibfield{author}{\bibinfo{person}{Yanan Sui}, \bibinfo{person}{Alkis Gotovos}, \bibinfo{person}{Joel Burdick}, {and} \bibinfo{person}{Andreas Krause}.} \bibinfo{year}{2015}\natexlab{}.
\newblock \showarticletitle{Safe exploration for optimization with Gaussian processes}. In \bibinfo{booktitle}{\emph{International conference on machine learning}}. PMLR, \bibinfo{pages}{997--1005}.
\newblock


\bibitem[\protect\citeauthoryear{Sui, Zhuang, Burdick, and Yue}{Sui et~al\mbox{.}}{2018}]%
        {sui2018stagewise}
\bibfield{author}{\bibinfo{person}{Yanan Sui}, \bibinfo{person}{Vincent Zhuang}, \bibinfo{person}{Joel Burdick}, {and} \bibinfo{person}{Yisong Yue}.} \bibinfo{year}{2018}\natexlab{}.
\newblock \showarticletitle{Stagewise safe Bayesian optimization with Gaussian processes}. In \bibinfo{booktitle}{\emph{International conference on machine learning}}. PMLR, \bibinfo{pages}{4781--4789}.
\newblock


\bibitem[\protect\citeauthoryear{Wang, Balakrishnan, and Singh}{Wang et~al\mbox{.}}{2019}]%
        {wang2019optimization}
\bibfield{author}{\bibinfo{person}{Yining Wang}, \bibinfo{person}{Sivaraman Balakrishnan}, {and} \bibinfo{person}{Aarti Singh}.} \bibinfo{year}{2019}\natexlab{}.
\newblock \showarticletitle{Optimization of smooth functions with noisy observations: Local minimax rates}.
\newblock \bibinfo{journal}{\emph{IEEE Transactions on Information Theory}} \bibinfo{volume}{65}, \bibinfo{number}{11} (\bibinfo{year}{2019}), \bibinfo{pages}{7350--7366}.
\newblock


\bibitem[\protect\citeauthoryear{Williams and Rasmussen}{Williams and Rasmussen}{2006}]%
        {williams2006gaussian}
\bibfield{author}{\bibinfo{person}{Christopher~KI Williams} {and} \bibinfo{person}{Carl~Edward Rasmussen}.} \bibinfo{year}{2006}\natexlab{}.
\newblock \bibinfo{booktitle}{\emph{Gaussian processes for machine learning}}. Vol.~\bibinfo{volume}{2}.
\newblock \bibinfo{publisher}{MIT Press}.
\newblock


\bibitem[\protect\citeauthoryear{Yang, Emmerich, Deutz, and B{\"a}ck}{Yang et~al\mbox{.}}{2019}]%
        {yang2019efficient}
\bibfield{author}{\bibinfo{person}{Kaifeng Yang}, \bibinfo{person}{Michael Emmerich}, \bibinfo{person}{Andr{\'e} Deutz}, {and} \bibinfo{person}{Thomas B{\"a}ck}.} \bibinfo{year}{2019}\natexlab{}.
\newblock \showarticletitle{Efficient computation of expected hypervolume improvement using box decomposition algorithms}.
\newblock \bibinfo{journal}{\emph{Journal of Global Optimization}} \bibinfo{volume}{75}, \bibinfo{number}{1} (\bibinfo{year}{2019}), \bibinfo{pages}{3--34}.
\newblock


\bibitem[\protect\citeauthoryear{Zanette, Zhang, and Kochenderfer}{Zanette et~al\mbox{.}}{2018}]%
        {zanette2018robust}
\bibfield{author}{\bibinfo{person}{Andrea Zanette}, \bibinfo{person}{Junzi Zhang}, {and} \bibinfo{person}{Mykel~J Kochenderfer}.} \bibinfo{year}{2018}\natexlab{}.
\newblock \showarticletitle{Robust super-level set estimation using gaussian processes}. In \bibinfo{booktitle}{\emph{Joint European Conference on Machine Learning and Knowledge Discovery in Databases}}. Springer, \bibinfo{pages}{276--291}.
\newblock


\bibitem[\protect\citeauthoryear{Zhang and Golovin}{Zhang and Golovin}{2020}]%
        {zhang2020random}
\bibfield{author}{\bibinfo{person}{Richard Zhang} {and} \bibinfo{person}{Daniel Golovin}.} \bibinfo{year}{2020}\natexlab{}.
\newblock \showarticletitle{Random hypervolume scalarizations for provable multi-objective black box optimization}. In \bibinfo{booktitle}{\emph{International conference on machine learning}}. PMLR, \bibinfo{pages}{11096--11105}.
\newblock


\end{thebibliography}

\clearpage
\appendix

\section{Additional Analyses}
\subsection{Acquisition Approximation Error}\label{app:error_decomp}

\begin{figure*}[t]
    \centering
    \includegraphics[width=0.95\textwidth]{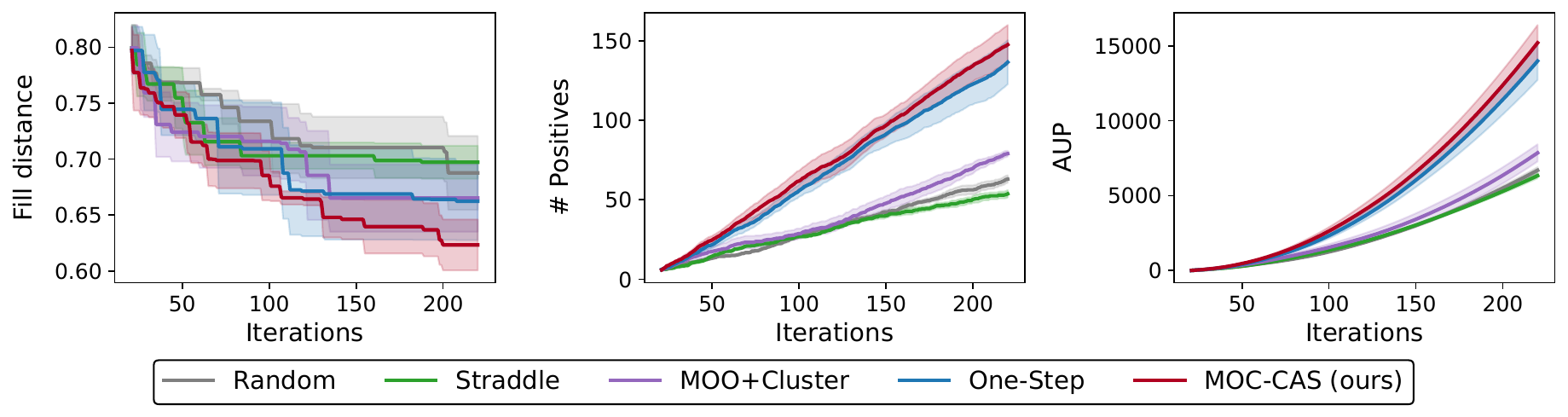}
    \caption{Quantitative comparison on the \textit{Cancer WRN} target dataset.}
    \label{fig:cancer_wrn}
    \Description{Quantitative comparison on the cancer WRN target dataset. The figure contains multiple line plots over iterations comparing methods with three metrics: fill distance, cumulative number of feasible discoveries, and the area under the positives curve.}
\end{figure*}

Let \(\alpha\) denote the true acquisition function and \(\tilde{\alpha}\) denote our approximated acquisition function. Our analysis shows that \(\forall x \in \mathcal{X}, t \in [T], |\alpha_{t-1}(x) - \tilde{\alpha}_{t-1}(x)| \le V_m(r)(\epsilon_1 + \epsilon_2)\), where \(V_m(r)\) denotes the volume of an \(m\)-dimensional ball of radius \(r\), \(\epsilon_1\) is the soft feasibility gate error, and \(\epsilon_2\) is the soft union error.

Here, \(\epsilon_1\) depends only on the feasibility margin, not \(t\). As the GP posterior sharpens and \textsc{MOC-CAS} prefers points that lie deeper inside the feasible orthant, this term decays exponentially with the distance from the thresholds. In contrast, \(\epsilon_2\) arises from summing the Gaussian overlap with all past outcomes, which introduces a linear factor in \(t\), but each summand is exponentially small whenever the candidate outcome is well separated from previously observed ones. Since MOC-CAS explicitly favors such outcome-space new points, the exponential decay in these overlaps dominates the linear factor, keeping \(\epsilon_2\) small even as \(t\) grows.

\subsection{Time Complexity of \textsc{MOC-CAS}}\label{app:complexity}

We provide our time complexity in this section. Let \(m\) denote the number of objectives, \(T\) denote total iterations, \(N = |\mathcal{X}|\) denote candidate set size, and \(d\)
denote input dimension.

\paragraph{Overall complexity.} With incremental GP updates and a finite candidate set, the overall time complexity of \textsc{MOC-CAS} is \(O(mT^3) + O(NmdT^2)\), with the \(O(NmdT^2)\) term typically dominating for large \(N\).

\paragraph{GP updates.} At round \(t\), each of the \(m\) GPs has \(t - 1\) data points. Using incremental Cholesky updates, one GP update costs \(O(t^2)\), so per round this is \(O(mt^2)\), and over \(T\) rounds:
\[
\sum_{t=1}^{T} O(mt^2) = O(mT^3).
\]

\paragraph{Acquisition maximization (finite \(\mathcal{X}\)).} At round \(t\), evaluating \(\tilde{\alpha}_{t-1}(x)\) for one candidate \(x\) requires: (i) GP prediction for all \(m\) objectives: \(O(mtd)\) with precomputed Cholesky factors, (ii) the soft feasibility gate and novelty term: \(O(m)\) and \(O(mt)\), respectively. The dominant cost is therefore \(O(mtd)\) per candidate. For exhaustive maximization over \(N\) candidates:
\[
\sum_{t=1}^{T} O(Nmtd) = O(NmdT^2).
\]

\section{Additional Experimental Details}\label{app:details}

\subsection{Settings of Datasets}

From DrugImprover, we then construct five per-molecule objectives from the given raw docking outputs using RDKit chemistry:
\begin{itemize}
    \item a docking objective derived from the raw OEDOCK scores (converted to “higher-is-better” and normalized per target),
    \item predicted solubility (ESOL from RDKit),
    \item synthetic accessibility (SA),
    \item RDKit QED which indicates drug-likeness, and
    \item a structural similarity\_to\_topK objective given by the maximum Tanimoto similarity to 1024 top-scoring ligands. 
\end{itemize}
All five objectives are scaled to $[0,1]$. For molecular representations, we develop SMILES–derived numeric features (RDKit descriptors) using RDKit \citep{rdkit, riniker2013fingerprints}. After the data preprocessing steps, the \textit{6T2W}, \textit{RTCB}, \textit{WRN}, and \textit{3CLPro} datasets have 119, 197, 193, and 194 SMILES–derived numeric features, respectively, that we use to conduct the experiments. We split all the $10^6$ molecules into 20 shards and featurize each  independently. Thus, all the experiments are conducted on four different drug–discovery tasks with five objectives per molecule (dock, solubility, SA, QED, and similarity-to-top\,$K$), where the per–dataset thresholds \(\tau\in\R^m\) define feasibility in the objective space. In our experiments, we set the per–dataset thresholds such that, for each dataset, the feasible set contains approximately 30\% of the full pool of 1 million molecules.

We model each objective using an independent GP and prefit kernel hyperparameters on a random subset of 200 samples. We then update the GP posteriors at each iteration with the cumulative data available up to that iteration. We run a budget of \(T{=}200\) iterations with a common warm start of \(n_{\mathrm{init}}{=}20\) uniformly drawn designs (220 iterations in total). Across \(4\) randomized trials, we reuse the same seeds and initial indices for all five methods in order to enable paired comparisons. In order to ensure scalability, all methods evaluate their acquisition function on a shared and light–weight shortlist developed each iteration by integrating optimistic per–objective prefiltering with random exploration under fixed caps. We break the ties by farthest distance in objective space from the previously observed outcomes.

\subsection{Additional Results on \textit{Cancer} Dataset}

In Figure \ref{fig:cancer_wrn} we show additional results on \textit{Cancer WRN} dataset, which shows similar performances of \textsc{MOC-CAS} algorithm as in the main paper.

\subsection{Additional Sample-Efficiency Metric: {T@50}}\label{app:t50}

Apart from the evaluation metrics used in the main paper, in order to quantify sample efficiency, we define a new evaluation metric, \(T@X = \min t : P(t) \ge X\), which is the number of iterations \(T\) needed to obtain \(X\) satisfactory samples. Following the same experimental setups in the main paper, we observe that MOC-CAS reaches 50 feasible samples significantly faster than all baselines. Table~\ref{tab:t50_6t2w} shows T@50 on \textit{Cancer 6T2W} dataset. Similar performances have been observed on other datasets too.

\begin{table}[t]
	\caption{\(T@50\) on \textit{Cancer 6T2W} dataset (smaller is better).}
	\label{tab:t50_6t2w}
	\begin{tabular}{ll}\toprule
		\textit{Method} & \textit{\(T@50\) (mean \(\pm\) std over 4 trials)} \\ \midrule
		\textsc{RANDOM} & \(171.2 \pm 14.6\) \\
		\textsc{ONE-STEP} & \(100.0 \pm 26.7\) \\
		\textsc{STRADDLE} & \(>220\) (not converge) \\
		\textsc{MOO+CLUSTER} & \(152.5 \pm 42.4\) \\
		\textbf{\textsc{MOC-CAS (ours)}} & \textbf{\(75.0 \pm 5.4\)} \\ \bottomrule
	\end{tabular}
\end{table}

\end{document}